\documentclass[conference]{IEEEtran}
\IEEEoverridecommandlockouts
\usepackage{cite}
\usepackage{amsmath,amssymb,amsfonts,amsthm}
\usepackage{algorithmic,comment}
\usepackage{graphicx}
\usepackage{textcomp}
\usepackage{xcolor}
\usepackage{comment}
\usepackage{graphicx}
\usepackage{textcomp}
\usepackage{xcolor,enumerate}
\usepackage{hyperref}

\usepackage{tikz}
\usepackage{pgfplots}

\usepackage{verbatim}

\def\BibTeX{{\rm B\kern-.05em{\sc i\kern-.025em b}\kern-.08em
		T\kern-.1667em\lower.7ex\hbox{E}\kern-.125emX}}
		
		\newcommand\copyrighttext{%
  \footnotesize \textcopyright 2020 IEEE. Personal use of this material is permitted.
  Permission from IEEE must be obtained for all other uses, in any current or future
  media, including reprinting/republishing this material for advertising or promotional
  purposes, creating new collective works, for resale or redistribution to servers or
  lists, or reuse of any copyrighted component of this work in other works. \href{https://doi.org/10.1109/IJCNN48605.2020.9207156}{DOI:10.1109/IJCNN48605.2020.9207156}}
\newcommand\copyrightnotice{%
\begin{tikzpicture}[remember picture,overlay]
\node[anchor=south,yshift=10pt] at (current page.south) {\fbox{\parbox{\dimexpr\textwidth-\fboxsep-\fboxrule\relax}{\copyrighttext}}};
\end{tikzpicture}%
}

\begin{document}
	
	\title{Language Inference with Multi-head Automata through Reinforcement Learning}
	
	\author{\IEEEauthorblockN{Alper Şekerci}
		\IEEEauthorblockA{\textit{Department of Computer Science} \\
			\textit{Özyeğin University}\\
			İstanbul, Turkey \\
			alper.sekerci@ozu.edu.tr}
		\and
		\IEEEauthorblockN{Özlem Salehi}
		\IEEEauthorblockA{\textit{Department of Computer Science} \\
			\textit{Özyeğin University}\\
			İstanbul, Turkey \\
			ozlem.koken@ozyegin.edu.tr}}
	
	\maketitle
	 \copyrightnotice 
	\begin{abstract}
		The purpose of this paper is to use reinforcement learning to model learning agents which can recognize formal languages. Agents are modeled as simple multi-head automaton, a new model of finite automaton that uses multiple heads, and six different languages are formulated as reinforcement learning problems. Two different algorithms are used for optimization. First algorithm is \textit{Q}-learning which trains gated recurrent units to learn optimal policies. The second one is genetic algorithm which searches for the optimal solution by using evolution-inspired operations. The results show that genetic algorithm performs better than \textit{Q}-learning algorithm in general but \textit{Q}-learning algorithm finds solutions faster for regular languages.
	\end{abstract}
	

	\section{Introduction}
	
	Grammatical inference is the process of learning a formal language from a set of labeled examples. It has various applications in the fields of pattern recognition, natural language processing, and computational biology. Its origins date back to the seminal work of Gold in 1960s \cite{GOLD1967447}. Since then, it has been investigated by many researchers including Fu \cite{fu1982syntactic}, Angluine and Smith \cite{inductive_inference}, Miclet \cite{gram_inference}.
	
	Considering the different approaches developed for grammatical inference, there has been a great interest in learning languages using recurrent neural networks (RNN). Some early examples include works of Elman \cite{ELMAN1990179} and Cleeremans et al. \cite{cleeremans} where first order RNNs are trained for regular language recognition. The problem is formulated as sequence prediction task, where the model is presented a single input symbol at each time step and predicts the next symbol. Following the work of Elman and Cleeremans et al., Giles et al. use second order RNNs to learn and extract finite automata for regular languages \cite{second_order_rnn}. Challenging harder languages, Das et al. \cite{learning_cfg} proposed an RNN model with an external stack to learn context-free languages. 
	
	An important line of research was opened by the study of long short-term memory (LSTM) \cite{lstm_ilk} networks in language recognition. Gers et al. \cite{gers_lstm} showed that LSTM networks can learn context-free and context-sensitive languages such as $a^nb^n$ and $a^nb^nc^n$. In 2018, Weiss et al. \cite{lstm_vs_gru} showed that LSTM is equivalent to a variant of multicounter automata \cite{counter_machines} and hence perform unbounded counting while recognizing languages like $a^nb^nc^n$ whereas Gated Recurrent Units (GRU) \cite{gru_ilk} can not, when worked under finite precision regime.\footnote{Note that RNNs with infinite precision are Turing complete in theory \cite{SIEGELMANN1995132}.} Another related work is due to Zaremba et al. \cite{rel_output} where the task is not to learn languages, but simple algorithms which can be carried on by a finite automaton working as a transducer and they use both supervised and reinforcement learning while training GRU and LSTM networks to learn finite automata accomplishing the task.

An alternative method for grammatical inference is the usage of evolutionary algorithms for inducing automata. Zhou et al. \cite{zhou_h_genetic} and Dupont \cite{genetic_search_dupont} use genetic algorithm \cite{genetic_ilk} to learn finite automata recognizing regular languages. Later on Lankhorst \cite{lankhorst_a_genetic_algorithm} and Huijsen \cite{genetic_gram_inference} apply genetics algorithm for the inference of context-free grammars and pushdown automata. Some more recent works on the subject include \cite{smart_state_label, evo_dfa, active_learning}.

In this paper, we introduce a new finite automaton model with multiple heads, namely simple multi-head automaton (SMA) and show that intelligent agents modeled as SMA can learn formal languages. The language recognition task is not defined as sequence prediction task as opposed to most of the studies from the literature but the automaton makes the decision of acceptance or rejection as a result of a sequential processing. Accordingly, we use reinforcement learning instead of supervised learning, expanding the previous work on the subject.  

Each language is formulated as an environment where agents can act on. At each timestep, an agent receives \textit{observation} from the environment and it performs an \textit{action} which either moves one of the heads on the tape or terminates the environment by accepting or rejecting the input string according to its policy. After each action, the agent receives a \textit{reward} and maximizing this reward leads to the correct and efficient decision on the input string.\par

Two different algorithms are implemented to optimize the policy of the agents while finding optimal SMA for various languages. The first algorithm is \textit{$Q$-learning}, where the policy of the agents are represented with GRUs and optimized by storing an experience buffer which is filled upon interacting with the environment. The second algorithm is \textit{genetic algorithm}, which does not improve the policy iteratively but instead tries to improve the whole population of agents by evaluating the \textit{fitness} of each agent and creating new individuals from the best agents with the hope of creating a better generation.\par
	
	To obtain results about the performance of each algorithm, 6 different languages are tested: 2 regular, 2 context-free and 2 non-context-free languages. Both the agents trained by $Q$-learning and genetic algorithm accomplished to recognize the regular languages $100\%$ correctly, but the agents trained with $Q$-learning achieved the results in a shorter time. For the other languages, genetic algorithm showed significantly better performance than $Q$-learning. Our results suggest that genetic algorithm deserves more attention in the area of grammatical inference.\par
	
	In Section \ref{sec: sma}, we define our new multi-head automaton model. Section \ref{sec: rl} describes the environment design and how an agent interacts with it. Section \ref{sec: q-agents} and \ref{sec: g-agents} contain information about the insights of $Q$-learning and genetic algorithm and further details about implementation. We present the results in Section \ref{sec: results} and conclude with Section \ref{sec: conc}.

	\newcommand{\startmarker}{\$}
	\newcommand{\finishmarker}{\#}
	
	\newcommand{\mysubsubsection}[1]{
		\vspace{0.1in}
		\subsubsection{#1}
	}
	
	\section{Simple Multi-head Finite Automata}\label{sec: sma}
	As the main purpose of this research is to model finite automata as learning agents for solving decision problems, a new model of multi-head finite automata is introduced with the motivation of reducing the parameter count that is required to be optimized during the learning process. 
	
	A \textit{simple multi-head automaton} (SMA) is a deterministic finite automaton that uses multiple heads.
	
Formally, a \textit{two-way simple $k$-head automaton} (2SMA($k$)) is a 9-tuple ($Q$, $q_0$, $F$, $\Sigma$, $\startmarker$, $\finishmarker$, $\delta$, $k$, $H$) where
	\begin{itemize}
		\item $Q$ is the set of states
		\item $q_0 \in Q$ is the initial state
		\item $F \subseteq Q$ is the set of accept states
		\item $\Sigma$ is the input alphabet.
		\item $\delta$ is transition function which maps $Q\times \tilde{\Sigma}$ into $Q$
		\item $k$ is the number of heads
		\item $H$ is the head assignment function which maps $Q$ into $\{\xleftarrow{},\;\circ,\;\xrightarrow{}\} \times \{\textup{head}_i \;|\; 0 \leq i < k\}$.
	\end{itemize}
	
	A machine is \textit{two-way} if the tape head can move right ($\rightarrow$), left ($\leftarrow$) and stay put ($\circ$). By restricting the head movements to the set $\{\circ,\;\xrightarrow{}\}$, we obtain a \textit{one-way simple $k$-head automaton} (1SMA($k$)).

	SMA uses a single finite input tape and the square with index 1 corresponds to the first symbol of the input string. Let $n$ denote the length of the input string. Then, the index 0 contains the start-marker {$\startmarker$} and the index $(n+1)$ contains the end-marker {$\finishmarker$}. Note that when the input string is empty, the index of the end-marker is 1.\par

	Initially, all heads start from the square with index 1 and the computation starts from the initial state. At each state, first the head and the direction that are assigned to the state are determined by the head assignment function $H$. After that, the head is moved 1 step in the assigned direction (no movement if $\circ$ is assigned) and then the symbol which the head is on is read. Note that the movement occurs before reading. Also, moving beyond start and end-markers is not allowed.\par
	
		After reading the symbol, SMA performs a transition using $\delta$ and enters into a new state. If there are no available transitions, the machine halts. If the machine halts in an accept state, the input string is accepted, and rejected otherwise. An SMA is said to \textit{recognize} a language \textit{L} if it accepts all and only the members of $L$.

		It is easy to see that 1SMA(1) is an ordinary deterministic finite automaton (DFA) and recognize exactly the class of regular languages. When compared to the classical multi-head finite automata (DFA($k$)) in which there are $k$ heads reading from an input tape simultaneously \cite{rosenberg_mfa,mfa}, 
		it turns out that the two models SMA($k$) and DFA($k$) are equivalent in terms of language recognition power.	The proof is omitted here. Note that the language recognition power of multi-head finite automata increase as the number of heads increase both for one-way and two-way models and  two-way models outperform one-way models for a constant number of heads \cite{mfa}.

	\section{Reinforcement Learning}\label{sec: rl}
	In this section, we will discuss how the components of the reinforcement learning algorithm are defined for the task of language recognition by simple multi-head automata.
	\begin{comment}
		In reinforcement learning, the purpose of a learning agent is to maximize the expected future discounted reward. The learning setup is as follows: There is an environment and an agent which acts on it. The environment provides the observation to the agent and with this observation, the agent decides on an action and applies it in the environment. As a result, the environment gives a reward to the agent.\par
		
		There is a wide range of reinforcement learning algorithms and many researchers achieved successful results. Using function approximators such as neural networks is a common approach. Nevertheless, there also exist tabular algorithms that only use arrays.
		
		\subsection{Environment \& Agent}
		An environment in a reinforcement learning setting has 4 responsibilities:
		\begin{enumerate}
			\item Preparing observation for the agent.
			\item Changing the environment state according to agent action.
			\item Terminating the environment when some condition is met.
			\item Giving positive or negative reward to the agent depending on its action.
		\end{enumerate}
				
	\noindent In this section, an environment for decision problems will be defined. The agent will represent an SMA.
	\end{comment}
	
	\subsection{Environment \& Agent}\label{sec: env}
	Let's start by describing the environment and the agent. The agent is a simple multi-head automaton. It can be one-way or two-way depending on the setting. 
	
	\subsubsection{Initial State}
	As it is mentioned while defining SMA, there is a finite input tape where the first square contains the start-marker and the last square contains end-marker. When the environment is reset, all heads of the SMA will be moved to square 1, which corresponds to the first symbol of the input string. According to the agent's actions, these heads will change their positions on the tape.
	
	\subsubsection{Observation}
	The transition function of an SMA dictates that only a single symbol can be read by a single head at a time. Note that the current state determines which head will be active and reading. Furthermore, a desired property for the observation is that the history of the observations should give all necessary information about the current state of the environment. 
	Thus, the observation contains only a single input symbol, index of the head by which the symbol is read and the direction in which the head moves.
	
	\subsubsection{Processing the Action}
	After receiving the observation, the agent will decide on its action. If the agent wants to terminate, which corresponds to the case where there is no valid transition in the current state, the agent can accept and terminate or reject and terminate. If it decides to continue, then it has to determine which head to move and its direction. Therefore, there are $(d \cdot h)$ possible head actions, where $d$ is the number of directions and $h$ is the number of heads.
	
	\subsubsection{Termination \& Reward}
	Theoretically, an SMA may never halt. Due to practical reasons, we put a limit on the maximum number of actions $N$ that the agent can perform during each episode. We set $N$ as $(2 \cdot M + 1) \cdot k + 1$ where $k$ is the number of heads and $M$ is the maximum length of the input string. This limit allows agent to move all heads to the end-marker and back to the start-marker before making a decision. Note that the maximum length of input strings is also limited because of practical reasons.\par
	
	The environment is terminated when the number of actions performed by the agent reaches $N$ or if the agent decides to terminate early. After each reset, a new input string is generated and it is determined whether it is a member string or not by a hand-crafted test function. When the environment is terminated, the agent receives a reward of +1 if it answers correctly, that is, accepts a member string or rejects a non-member string. It receives a reward of -1 if the answer is wrong, and no reward for actions without termination. There are 2 special cases for terminal rewards: If the agent answers wrong without reaching the end-marker, it is encouraged to read the whole input string before terminating to make sure that it has the correct answer and therefore it receives a reward of -10. If the agent waits until the very end of the episode to reject a string, it receives only a reward of 0.1 which discourages the agent from waiting too long if it is sure about the answer.
	
	\section{$Q$-Agents}\label{sec: q-agents}
One way to optimize the policy of an agent is $Q$-learning. The agents trained using $Q$-learning algorithm will be called \textit{$Q$-agents}. In this section, we will describe the details of the $Q$-learning algorithm.
	
	\subsection{Deep $Q$-Learning}
	
	$Q$-value is a measure of how good is it to perform action $a$ in state $q$. The function $Q(q, a)$ is defined as 
	$$Q(q, a) = R(q, a) + \gamma \cdot V(q_{next})$$
	where $R$ is the immediate reward received by performing the action $a$ in state $q$, $\gamma$ is the discount factor which makes the rewards that are received sooner more favorable and $q_{next}$ is the next state the agent moves in after performing the action. 
	
	According to the Bellman Equation \cite{atari}, the value $V$ of a state $q$ is simply the maximum $Q$-value the agent can get in a given state by performing any action.

$$V(q) = max_a(Q(q, a))$$

To learn the optimal $Q$ function, it is possible to use either arrays or a neural network to approximate the function. In deep $Q$-learning, a deep neural network is used to approximate the $Q$ function. 

\subsection{GRU vs. LSTM}

In order to provide internal memory for $Q$-learning agents, \textit{gated recurrent units (GRU)} are used in this paper. GRU is a simpler alternative to \textit{long short-term memory (LSTM)}. It is known that LSTM is more successful in language recognition as it can perform unbounded counting \cite{lstm_vs_gru}. The reason why GRU is preferred over LSTM in this paper is to test a new multi-head automaton model focusing on the effect of multiple heads and ability of moving left. As a recurrent neural network model which can perform counting can easily learn languages like $a^nb^n$ or $a^nb^nc^n$ using a single-head and moving in a single direction, GRUs which cannot perform counting suit better the purpose of this paper.

	\subsection{Modeling SMA with Neural Networks}
	The automata defined in this paper have discrete states. However, a continuous state space is needed to train neural networks using gradient descent method.
	
	In the discrete case, each state can be represented by an integer and a boolean lookup table can be used to determine which states are accepting.
	
	In the continuous case, a state can be represented by a real vector. Thus, the transition function $\delta$ takes as input no longer an integer but a vector and the one-hot encoding of an input symbol, and outputs a vector. Instead of a lookup table for determining the acceptance of a state, a new function $A$ maps the state vector to a 3 dimensional stochastic vector, representing a probability distribution over three types of states:
	\begin{enumerate}[i.]
		\item rejecting but not halting,
		\item rejecting and halting,
		\item accepting and halting.
	\end{enumerate}
	
	\noindent So, the function $A$ randomly samples one of these types according to the probability distribution and assigns it as the type of the input state.\par
	
	Similarly for the head-movement, a function $M$ maps an input state vector to a $2k$-dimensional and $3k$-dimensional stochastic vector for 1SMA($k$) and 2SMA($k$), respectively. Then, the function $M$ randomly samples the action for the head movement and assigns it to the input state.

\subsection{Implementation}

	As explained in Section \ref{sec: env}, the number of possible actions $A$ for the agent is ${2 + (d \cdot k)}$, where $d$ is the number of directions and $k$ is the number of heads. Therefore, there is a $Q$-network that takes the current internal state, which is the output of the last recurrent unit, as input. Then, there are fully-connected hidden layers, the layer count and the number of neurons in each layer are hyperparameters. The final layer is the output layer with dimension $A$.\par
	
We use two methods to improve the stability and convergence of deep $Q$-networks. First, it is possible to store \textit{experiences} in a buffer \cite{atari}. An experience is a tuple ($s_t$, $action$, $reward$, $s_{t+1}$, $done$) where $s_t$ is the observation before performing the action, $s_{t+1}$ is the observation after performing the action and $done$ represents if the environment is terminated after the action. While training, a batch of experiences is sampled uniformly from this buffer.\par
	
	The second method is using fixed target network \cite{target_net}. $Q$-learning uses the estimation for the next state while updating $Q$-value of the current state. With this method, the estimation will not be taken from the network which is currently being trained but from a fixed $Q$-network and the weights of the trained $Q$-network is copied onto the target network periodically.\par
	
	During training, an agent plays many episodes to fill up the experience buffer. After the buffer is fulled, the neural network is optimized using the data in the buffer. At the start of each episode, the input string on the tape is changed. Thus, the experience buffer contains different strings with different lengths, which helps the agent to generalize better.\par
	
	Moreover, $Q$-learning agents use $\epsilon$-greedy exploration. That is, with $\epsilon$ probability an agent chooses a random action and with $(1-\epsilon)$ probability it chooses the best action. This hyperparameter handles the exploration-exploitation trade-off: exploration is for trying different actions to achieve better rewards and exploitation is for using the agent's current knowledge to maximize the rewards.

	\section{$G$-Agents}\label{sec: g-agents}
	Another approach for policy optimization in a reinforcement learning problem is using genetic algorithm. Agents trained with genetic algorithm will be called \textit{$G$-agents}.
	
	\subsection{Genetic Algorithm}
	\textit{Genetic algorithm} is a black-box optimization technique which uses operations inspired by biological evolution \cite{genetic}. A population of individuals is randomly initialized and each individual corresponds to a chromosome which is a chain consisting of genes. There exists a fitness function which takes a chromosome as input and returns its fitness value, that is, the performance measure of the chromosome for the given problem. Genetic algorithm works by improving the initial population at each generation.\par
	
	 In this approach, each SMA is represented with a chromosome and its performance is evaluated with a fitness function. Then, at each iteration a collection of chromosomes is improved by eliminating bad solutions and creating new chromosomes using the good solutions, with the aim of finding the most optimal automaton recognizing the language trained for.

	\begin{comment}
	 The algorithm is as follows:
	\begin{enumerate}[i.]
		\item Initialize a population with $N$ individuals randomly.
		\item\label{compute_fitness} Compute fitness value for each individual using $F$.
		\item Order the individuals according to their fitness.
		\item Remove the worst half of the population.
		\item For each removed individual, create a new individual from 2 parents chosen randomly from the other half.
		\item Repeat step \ref{compute_fitness} until convergence.
	\end{enumerate}
	
	Also, the algorithm to create a new individual (chromosome) from 2 parents is as follows:
	\begin{enumerate}[i.]
		\item Let $L$ be the length of a chromosome (all chromosomes have same length).
		\item Choose a cutting index $C$ randomly on the chromosome, in the range [1, L).
		\item Start performing the cross-over operation: copy the genes of $parent_1$ between indices [0, C) onto the new chromosome named \textit{child}.
		\item Copy the genes of $parent_2$ between indices [C, L) onto child.
		\item Choose the number of genes $M$ that will be mutated randomly.
		\item If M is zero, terminate.
		\item\label{start_mut} Choose a random index $i$.
		\item Replace the gene of index $i$ with a random gene.
		\item Repeat step \ref{start_mut} until the number of mutated genes becomes M.
	\end{enumerate}
	
	\end{comment}

	\subsection{Representation of SMA with Chromosome}
	To apply genetic algorithm, we need to represent an SMA with a string of integers making up a chromosome. The individual integers are called the \textit{genes}.
	
	Let $n$ be the number of states in the SMA and let $|\tilde{\Sigma}|=m$ where $\tilde{\Sigma}=\Sigma \cup \{\startmarker , \finishmarker\}$. There might be a transition between any pair of two states with any one of the $m$ symbols as its label. For each state, each possible transition is represented with a gene $g_{\xrightarrow{}}$ which holds the information about the target state of the transition. The range of each gene is [0, $n$], where 0 means that there is no transition and the remaining integers are the indices of the states. For each state, $m$ genes are required to represent all possible outgoing transitions from the state for each symbol.\par
	
	Moreover, head assignment function which assigns the head and the direction for each state is stored with a single gene $g_k$ in the range [0, $d \cdot k$), where $d$ is the number of directions and $k$ is the number of heads. Lastly, for each state a single gene $g_a$ in the range [0, 1] is required to store whether it is an accept state or not. As a result, a chromosome is a sequence of genes

$$(m \cdot g_{\xrightarrow{}}) \cdot n + g_k \cdot n + g_a \cdot n$$
where multiply represents duplication and plus represents concatenation.
	
	\subsection{Fitness}
	Initially, a training set is formed with $N$ strings which are generated randomly by the environment. This training set is used for computing the fitness value of an individual.\par
	
	Each individual is tested for $N$ different episodes, which contain the input strings on the tape chosen from the training set. For each episode, the total episode reward is stored and the sum of all episode rewards is used as the fitness value. Similar to $Q$-learning algorithm, the rewards are multiplied by a discount factor $\gamma$ to make sooner rewards more favorable.\par
	
	Moreover, when the best individual in a generation achieves 100\% correct prediction rate, a new training set is formed and all fitness values are recomputed so that if there exist some strings that are not accepted even by the best individual, the individual can improve itself further.
	
	\vspace{0.1in}
	
	\section{Results}\label{sec: results}
	
	\subsection{Languages}
	During training, agents are taught 6 different languages:
	\begin{enumerate}[i.]
		\item $L_1 = \{\;0w1 \;|\; w \in \{0, 1\}^*\}$
		\item $L_2 = \{\;w \;|\; w \in \{0, 1\}^*$ and length of $w$ is even $\}$
		\item $L_3 = \{\;a^nb^n \;|\; n \geq 0\}$
		\item $L_4 = \{\;w \;|\; w \in \{0, 1\}^*$ and $w$ is palindrome $\}$
		\item $L_5 = \{\;a^nb^nc^n \;|\; n \geq 0\}$
		\item $L_6 = \{\;ww \;|\; w \in \{0, 1\}^*\}$
	\end{enumerate}
	\vspace{0.05in}
	Note that $L_1$ and $L_2$ are regular and both can be recognized by 1SMA(1). $L_3$ and $L_4$ are non-regular but context-free and $L_5$ and $L_6$ are non-context-free languages. $L_3$ can be recognized by a 1SMA(2) but for $L_4$, 2SMA(2) is needed as the tape head should be able to move to both directions. $L_5$ is recognized by a 1SMA(2) and $L_6$ is recognized either by a 1SMA(3) or 2SMA(2) but not with a 1SMA(1).  
	
All languages except $L_4$ are trained on a 1SMA. $L_1$ and $L_2$ are trained with a single head, $L_4$ with 2 heads and finally $L_5$ and $L_6$ are trained with 3 heads. Note that for $L_5$ an extra head is added to test whether the algorithms can optimize and use less heads.
	
	\subsection{Hyperparameters}
	
		The hyperparameters for $Q$-learning are given below:
	\begin{itemize}
		\item The output size of a recurrent unit is 32.
		\item The discount factor $\gamma$ is 0.999.
		\item The $Q$-network which takes the output of the last recurrent unit as input has 1 hidden layer with 32 neurons that use $arctan$ activation function.
		\item The experience buffer size is 25000.
		\item $\epsilon$ for exploration is 0.05.
	\end{itemize}
	
	The hyperparameters for genetic algorithm are as follows:
	\begin{itemize}
		\item The population size is 100.
		\item The state size of SMA is 32.
		\item The chromosome length $C$ of an individual is ${(m + 2) \cdot n}$, where $m$ is the number of symbols and $n$ is the number of states.
		\item The maximum number of mutations is 3 for regular languages, $(C / 20)$ for other languages.
		\item The discount factor $\gamma$ is 0.999.
		\item The training set size is 1000.
	\end{itemize}
	\vspace{0.05in}
	
	\subsection{Discussion}
		Fig. \ref{rew_table} shows the performance of different algorithms for different languages. First column is the model name, second column is the average reward, third column is the correct prediction rate and the fourth column is the average episode length. The data is collected by running the algorithms for 10000 episodes, that is, for 10000 different input strings. Note that the maximum length of the input strings is set to 20, because of practical reasons mentioned before.\par

	The number in the model name represents the language the agent is taught and the rightmost letter represents the algorithm that the agent uses. $R$ is the random algorithm that performs a random action at each step, $Q$ and $G$ represent the $Q$-agent and $G$-agent respectively.
	
	An algorithm is commonly evaluated according to 3 criteria: correctness, memory usage and running time. The solutions found by $Q$-agent and $G$-agent can be also evaluated similarly. In the results, correct prediction rate shows the \textit{correctness} of the solution and the average episode length shows the \textit{running time}. Note that optimizing the memory usage is not a concern in this paper.

	\begin{figure}[htbp]
		\begin{center}
			\begin{tabular}{ |c|c|c|c| } 
				\hline
				Model & Avg. Reward & Pred. Rate & Avg. Ep. Length \\
				\hline\hline
				$L_1R$ & -4.193 & 0.498 & 2.0 \\\hline
				$L_1Q$ & 1.000 & 1.000 & 8.6 \\\hline
				$L_1G$ & 0.988 & 1.000 & 12.8 \\\hline
				\hline
				$L_2R$ & -4.151 & 0.504 & 2.0 \\\hline
				$L_2Q$ & 0.995 & 1.000 & 12.8 \\\hline
				$L_2G$ & 0.984 & 1.000 & 16.8 \\\hline
				\hline
				$L_3R$ & -4.156 & 0.498 & 3.0 \\\hline
				$L_3Q$ & -1.480 & 0.731 & 4.2 \\\hline
				$L_3G$ & 0.989 & 1.000 & 11.1 \\\hline
				\hline
				$L_4R$ & -4.027 & 0.511 & 4.0 \\\hline
				$L_4Q$ & -3.208 & 0.457 & 8.8 \\\hline
				$L_4G$ & 0.724 & 0.866 & 14.5 \\\hline
				\hline
				$L_5R$ & -4.089 & 0.499 & 4.0 \\\hline
				$L_5Q$ & -0.341 & 0.329 & 13.2 \\\hline
				$L_5G$ & 0.991 & 0.999 & 10.1 \\\hline % pred_rate: 0.99997
				\hline
				$L_6R$ & -4.172 & 0.490 & 4.0 \\\hline
				$L_6Q$ & -0.672 & 0.445 & 10.6 \\\hline
				$L_6G$ & 0.795 & 0.903 & 17.3 \\\hline
			\end{tabular}
		\end{center}
		\caption{Performance of models for different languages.}
		\label{rew_table}
	\end{figure}
		The table in Fig. \ref{head_table} further supports this result. This table provides statistics about head movement in different solutions. First column is the model name which represents the same models as in Fig. \ref{rew_table}, next three columns show the usage of each head compared to others and the last three columns show which direction the heads are moved mostly.\par
	\begin{figure}[htbp]
		\begin{center}
			\begin{tabular}{ |c|c|c|c|c|c|c| } 
				\hline
				Model & $h_1$ & $h_2$ & $h_3$ & $\xleftarrow{}$ & $\circ$ & $\xrightarrow{}{}$\\
				\hline\hline
				$L_1Q$ & 1.00 &  &  &  & 0 & 1.00 \\\hline
				$L_1G$ & 1.00 &  &  &  & 0.25 & 0.75 \\\hline
				\hline
				$L_2Q$ & 1.00 &  &  &  & 0.10 & 0.90 \\\hline
				$L_2G$ & 1.00 &  &  &  & 0.25 & 0.75 \\\hline
				\hline
				$L_3Q$ & 0.71 & 0.29 &  &  & 0.00 & 1.00 \\\hline
				$L_3G$ & 0.65 & 0.35 &  &  & 0.15 & 0.85 \\\hline
				\hline
				$L_4Q$ & 0.17 & 0.83 &  & 0.09 & 0.02 & 0.89 \\\hline
				$L_4G$ & 0.79 & 0.21 &  & 0.07 & 0.05 & 0.88 \\\hline
				\hline
				$L_5Q$ & 0.04 & 0.61 & 0.35 &  & 0.06 & 0.94 \\\hline
				$L_5G$ & 0.53 & 0.00 & 0.47 &  & 0.08 & 0.92 \\\hline
				\hline
				$L_6Q$ & 0.91 & 0.09 & 0.00 &  & 0.09 & 0.91 \\\hline
				$L_6G$ & 0.36 & 0.00 & 0.64 &  & 0.04 & 0.96 \\\hline
			\end{tabular}
		\end{center}
		\caption{Statistics about head movement.}
		\label{head_table}
	\end{figure}

		\subsubsection{Random Algorithm}

		The result of the random algorithm is included in order to better assess the performance of the other two algorithms. A random agent has no knowledge of the environment and it does not change its policy according to the state it is currently in. Any well designed algorithm is expected to perform better than the random algorithm.\par
	
	As expected, the random agents for all languages performed the worst. Note that the correct prediction rates for random agents are approximately 0.5, which means they made the correct decision for half of the strings. This is expected as with probability 0.5, the input string is chosen from the language whereas with probability 0.5 it is generated randomly.

\subsubsection{Regular Languages}
	
 For the regular languages $L_1$ and $L_2$, both $Q$-agent and $G$-agent achieved $100\%$ correct prediction. For the $Q$-agent, the average reward is higher and the average episode length is smaller in each case which shows that the $Q$-agent have learned more efficient solutions for these languages.\par
 
\begin{figure}[htbp]
		\begin{tikzpicture}
		\begin{axis}[
		title={Prediction rates of $Q$-learning algorithm for $L_1$ and $L_2$.},
		xlabel={Timesteps (1k)},
		ylabel={Avg. Correct Prediction Rate},
		xmin=0, xmax=1625,
		ymin=0.35, ymax=1.0,
		xtick={0,200,400,600,800,1000,1300,1625},
		ytick={0.35,0.45,0.50,0.75,0.80,0.90,0.95,1.0},
		legend pos=north east,
		ymajorgrids=true,
		grid style=dashed,
		]
		
		\addplot[
		color=black,
		mark=,
		]
		table {data/q_learning/L1q_rew.txt};
		
		\addplot[
		color=black!70,
		mark=,
		dashed,
		]
		table {data/q_learning/L2q_rew.txt};
		
		\legend{$L_1$\\$L_2$\\}
		
		\end{axis}
		\end{tikzpicture}
		\caption{The change of the average correct prediction rates for the regular languages during training with $Q$-learning algorithm.}
		\label{q_rew_graph}
	\end{figure}
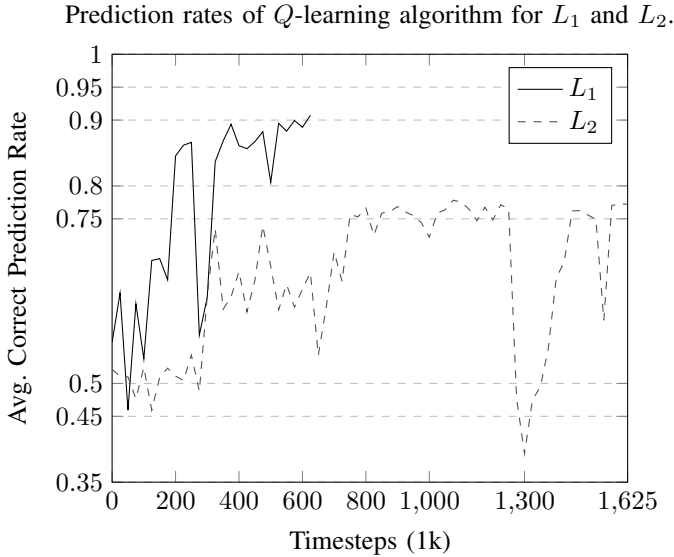

 Figure \ref{genetic_graph} and  Figure \ref{q_rew_graph} display prediction rates during training for $G$-agents and $Q$-agents respectively. Note that during training, the average correct prediction rates for $Q$-learning algorithm can be lower as the agents perform random actions with $\epsilon$ probability due to $\epsilon$-greedy exploration.\par

	\begin{figure}[htbp]
		\begin{tikzpicture}
		\begin{axis}[
		title={Prediction rates of genetic algorithm for $L_1$ and $L_2$.},
		xlabel={Generation},
		ylabel={Correct Prediction Rate},
		xmin=0, xmax=90,
		ymin=0.45, ymax=1,
		xtick={0,10,20,30,40,50,60,70,80,90},
		ytick={0.45,0.50,0.75,0.80,0.90,0.95,1.0},
		legend pos=south east,
		ymajorgrids=true,
		grid style=dashed,
		]
		
		\addplot[
		color=black,
		mark=,
		]
		table {data/genetic/L1g.txt};
		
		\addplot[
		color=black!70,
		mark=,
		dashed,
		]
		table {data/genetic/L2g.txt};
		
		\legend{$L_1$\\$L_2$\\}
		
		\end{axis}
		\end{tikzpicture}
		\caption{Correct prediction rates of the best individuals in each generation during the training of regular languages with genetic algorithm.}
		\label{genetic_graph}
	\end{figure}
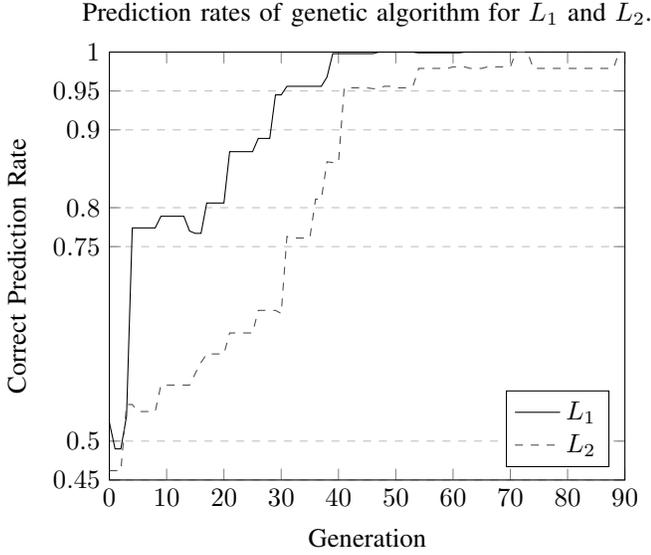
	
  As the head is allowed to stay or move in both directions while processing the input string, moving the heads efficiently reduces the time to reach the answer. Since $L_1$ and $L_2$ are regular, by definition they can be recognized by a real-time DFA in which the head always moves right. Looking at Figure \ref{head_table}, we see that $Q$-agent always moved the head right for $L_1$  whereas the $G$-agent stayed at the same position for $25\%$ of the steps.  This explains how the $Q$-agent can terminate earlier for $L_1$.\par

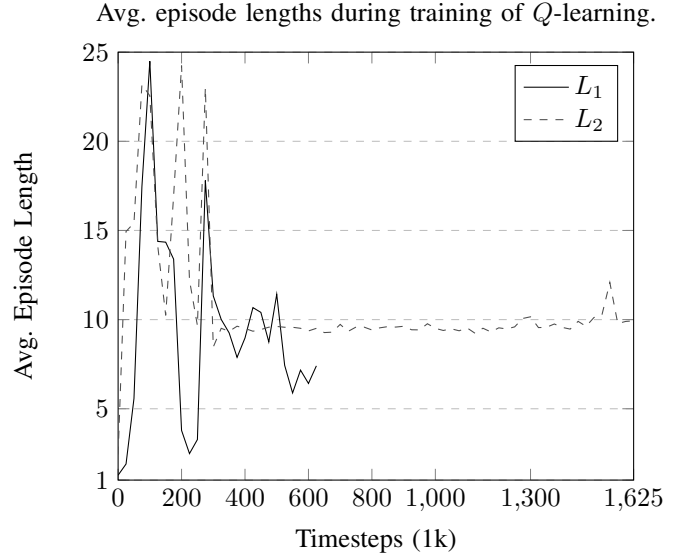
\begin{figure}[htbp]
		\begin{tikzpicture}
		\begin{axis}[
		title={Avg. episode lengths during training of $Q$-learning.},
		xlabel={Timesteps (1k)},
		ylabel={Avg. Episode Length},
		xmin=0, xmax=1625,
		ymin=1, ymax=25,
		xtick={0,200,400,600,800,1000,1300,1625},
		ytick={1, 5, 10, 15, 20, 25},
		legend pos=north east,
		ymajorgrids=true,
		grid style=dashed,
		]
		
		\addplot[
		color=black,
		mark=,
		]
		table {data/q_learning/L1q_length.txt};
		
		\addplot[
		color=black!70,
		mark=,
		dashed,
		]
		table {data/q_learning/L2q_length.txt};
		
		\legend{$L_1$\\$L_2$\\}
		
		\end{axis}
		\end{tikzpicture}
		\caption{The change of the average episode lengths for the regular languages during training with $Q$-learning algorithm.}
		\label{q_length_graph}
	\end{figure}

Early termination is another factor which reduces the solution time. For instance, in $L_1$ there is no string that starts with a 1, and therefore it is possible to terminate immediately if the automaton reads 1 at the beginning. However, in $L_2$ the automaton has to read all the input string to check whether the length of string is even or not. In fact the results show that the agents took less time to reach the answer for $L_1$. Figure \ref{q_length_graph} displays the average episode length during training of the $Q$-agents for both languages.

	\subsubsection{Nonregular Languages}

Even though $Q$-agent performed better for regular languages, it was not successful for the remaining languages. Nevertheless, for all languages it gained more reward than the random agent and it is possible to say that the agents managed to find sub-optimal solutions. However, this does not necessarily imply higher correct prediction rates. $Q$-agent has higher correct prediction rate than random agent only for $L_3$. \par
	
The reward function punishes heavily the agents which answer wrong without reading the whole input string. So, for an agent which can not find the correct answer, it is better to reach the end of the string first before terminating. This is how the $Q$-agents may have gained higher rewards than random agents.\par
	
	On the other hand, $G$-agents performed significantly better for non-regular languages. For $L_3$ $G$-agent achieved $100\%$ correct prediction rate and for $L_5$ it rarely answers wrong. There is an important detail in Fig. \ref{head_table} for $L_5G$. Even though there are 3 heads, the second head is not used at all which is expected for an efficient agent as theoretically $L_5$ can be recognized by a 2SMA(2).
	
	Furthermore, there is a critical observation about the input generation algorithm. As discussed earlier, half of the time the input string is generated randomly and half of the time it is chosen from the language, and the maximum length of the generated strings is 20. So, if agents can understand if a string is randomly-generated or not, then they can easily understand whether it is a member string or not.\par
	
	There is an important difference between $L_5$, and $L_4$ and $L_6$. In $L_4$ and $L_6$, first half of the string can actually be random and only the second half must obey some format regarding the first half. Therefore, it is possible to say that strings that are in $L_4$ and $L_6$ look more random than $L_5$. For instance, if a string starts with 5 $a$'s and continues with 5 $b$'s, then it already looks like a very organized string, thus an agent may assume the rest of the string will contain 5 $c$'s. So, at that step the agent may guess that the string is in the language and most of the time makes a correct guess.\par
	
	This can also explain why $G$-agents performed the worst in $L_4$ and $L_6$. As determining if the input string is random or not is harder for these languages, learning these languages is harder for the agents.

	\section{Conclusion}\label{sec: conc}
	In this paper, two different algorithms are analyzed and tested for training simple multi-head automata to recognize several decision problems. According to the results, genetic algorithm performed better overall.\par
	
	\subsection{$Q$-Learning vs. Genetic}
	Since $Q$-learning algorithm uses neural networks and applies gradient descent to optimize weights, it involves calculation with continuous values and calculus. On the other hand, genetic algorithm involves integers and uses evolution-inspired operations for optimization.\par
	
	Running a neural network is more costly while in genetic algorithm, integer arrays are used for simulating the automata making it faster. Additional cost of the neural networks could be justified with higher correct prediction rates whereas this is not the case. In fact, agents that use neural networks only learned to recognize regular languages. Thus, it is possible to say that the genetic algorithm turned out to be more effective and more efficient.\par
	
	\subsection{Future Work}
	In this paper, it is shown that a state of an automaton can be represented with continuous values and the transition function can map a state vector into another. This can also be done for the alphabet. Currently, a head reads a symbol which is a member of the finite set named alphabet. However, a symbol can actually be a real number.\par
	
	Continuous symbols are not necessarily useful for decision problems but can be useful when there is an output tape. A transducer automaton can write symbols on an output tape \cite{transducer}. In future work, a new transducer automaton model can be defined which optionally uses continuous states or continuous input/output symbols to learn different algorithms.\par
	
	Furthermore, as genetic algorithm looks promising, more advanced algorithms for population management and different methods for creating new individuals can be investigated. Also, testing with harder languages which require more than 3 heads and changing the input generation algorithm in a way that it tries to find corner cases for agents which they fail can improve the effectiveness of training and thus help the agents to generalize better.
	
	\bibliographystyle{IEEEtran}
	\bibliography{references.bib}{}

\begin{thebibliography}{10}
\providecommand{\url}[1]{#1}
\csname url@samestyle\endcsname
\providecommand{\newblock}{\relax}
\providecommand{\bibinfo}[2]{#2}
\providecommand{\BIBentrySTDinterwordspacing}{\spaceskip=0pt\relax}
\providecommand{\BIBentryALTinterwordstretchfactor}{4}
\providecommand{\BIBentryALTinterwordspacing}{\spaceskip=\fontdimen2\font plus
\BIBentryALTinterwordstretchfactor\fontdimen3\font minus
  \fontdimen4\font\relax}
\providecommand{\BIBforeignlanguage}[2]{{%
\expandafter\ifx\csname l@#1\endcsname\relax
\typeout{** WARNING: IEEEtran.bst: No hyphenation pattern has been}%
\typeout{** loaded for the language `#1'. Using the pattern for}%
\typeout{** the default language instead.}%
\else
\language=\csname l@#1\endcsname
\fi
#2}}
\providecommand{\BIBdecl}{\relax}
\BIBdecl

\bibitem{GOLD1967447}
E.~M. Gold, ``Language identification in the limit,'' \emph{Information and
  Control}, vol.~10, no.~5, pp. 447 -- 474, 1967.

\bibitem{fu1982syntactic}
K.~Fu, \emph{Syntactic Pattern Recognition and Applications}, ser.
  Prentice-Hall Advanced Reference Series: Computer Science.\hskip 1em plus
  0.5em minus 0.4em\relax Prentice-Hall, 1982.

\bibitem{inductive_inference}
D.~Angluin and C.~H. Smith, ``Inductive inference: Theory and methods,''
  \emph{ACM Comput. Surv.}, vol.~15, no.~3, pp. 237--269, Sep. 1983.

\bibitem{gram_inference}
L.~Miclet, ``Grammatical inference,'' in \emph{Syntactic and Structural Pattern
  Recognition—Theory and Applications}.\hskip 1em plus 0.5em minus
  0.4em\relax World Scientific, 1990, pp. 237--290.

\bibitem{ELMAN1990179}
J.~L. Elman, ``Finding structure in time,'' \emph{Cognitive Science}, vol.~14,
  no.~2, pp. 179 -- 211, 1990.

\bibitem{cleeremans}
A.~Cleeremans, D.~Servan-Schreiber, and J.~Mcclelland, ``Finite state automata
  and simple recurrent networks,'' \emph{Neural Computation - NECO}, vol.~1,
  pp. 372--381, Sep. 1989.

\bibitem{second_order_rnn}
C.~L. {Giles}, C.~B. {Miller}, D.~{Chen}, H.~H. {Chen}, G.~Z. {Sun}, and Y.~C.
  {Lee}, ``Learning and extracting finite state automata with second-order
  recurrent neural networks,'' \emph{Neural Computation}, vol.~4, no.~3, pp.
  393--405, May 1992.

\bibitem{learning_cfg}
S.~Das, C.~L. Giles, and G.~Sun, ``Learning context-free grammars: Capabilities
  and limitations of a recurrent neural network with an external stack
  memory,'' in \emph{Proceedings of The Fourteenth Annual Conference of
  Cognitive Science Society. Indiana University}, 1992, p.~14.

\bibitem{lstm_ilk}
S.~Hochreiter and J.~Schmidhuber, ``Long short-term memory,'' \emph{Neural
  Comput.}, vol.~9, no.~8, pp. 1735--1780, Nov. 1997.

\bibitem{gers_lstm}
F.~A. {Gers} and E.~{Schmidhuber}, ``Lstm recurrent networks learn simple
  context-free and context-sensitive languages,'' \emph{IEEE Transactions on
  Neural Networks}, vol.~12, no.~6, pp. 1333--1340, Nov. 2001.

\bibitem{lstm_vs_gru}
G.~Weiss, Y.~Goldberg, and E.~Yahav, ``On the practical computational power of
  finite precision {RNNs} for language recognition,'' in \emph{Proceedings of
  the 56th Annual Meeting of the Association for Computational Linguistics,
  {ACL} 2018, Melbourne, Australia, July 15-20, 2018, Volume 2: Short Papers},
  I.~Gurevych and Y.~Miyao, Eds.\hskip 1em plus 0.5em minus 0.4em\relax
  Association for Computational Linguistics, 2018, pp. 740--745.

\bibitem{counter_machines}
P.~Fischer, A.~Meyer, and A.~Rosenberg, ``Counter machines and counter
  languages,'' \emph{Theory of Computing Systems}, vol.~2, pp. 265--283, Sep.
  1968.

\bibitem{gru_ilk}
K.~Cho \emph{et~al.}, ``Learning phrase representations using rnn
  encoder-decoder for statistical machine translation,'' \emph{arXiv preprint
  arXiv:1406.1078}, 2014.

\bibitem{SIEGELMANN1995132}
H.~Siegelmann and E.~Sontag, ``On the computational power of neural nets,''
  \emph{Journal of Computer and System Sciences}, vol.~50, no.~1, pp. 132 --
  150, 1995.

\bibitem{rel_output}
W.~Zaremba, T.~Mikolov, A.~Joulin, and R.~Fergus, ``Learning simple algorithms
  from examples,'' in \emph{International Conference on Machine Learning},
  2016, pp. 421--429.

\bibitem{zhou_h_genetic}
H.~Zhou and J.~J. Grefenstette, ``Induction of finite automata by genetic
  algorithms,'' in \emph{Proceedings of the 1986 IEEE International Conference
  on Systems, Man and Cybernetics}, 1986, pp. 170--174.

\bibitem{genetic_search_dupont}
P.~Dupont, ``Regular grammatical inference from positive and negative samples
  by genetic search: the gig method,'' in \emph{Grammatical Inference and
  Applications}, R.~C. Carrasco and J.~Oncina, Eds.\hskip 1em plus 0.5em minus
  0.4em\relax Berlin, Heidelberg: Springer Berlin Heidelberg, 1994, pp.
  236--245.

\bibitem{genetic_ilk}
J.~H. Holland, \emph{Genetic Algorithms and Adaptation}.\hskip 1em plus 0.5em
  minus 0.4em\relax Boston, MA: Springer US, 1984, pp. 317--333.

\bibitem{lankhorst_a_genetic_algorithm}
M.~M. Lankhorst, \emph{Genetic algorithms in data analysis}.\hskip 1em plus
  0.5em minus 0.4em\relax Rijksuniversiteit Groningen, 1996.

\bibitem{genetic_gram_inference}
W.~Huijsen, ``Genetic grammatical inference,'' in \emph{CLIN IV: Papers from
  the Fourth CLIN Meeting}.\hskip 1em plus 0.5em minus 0.4em\relax Citeseer,
  1993, pp. 59--72.

\bibitem{smart_state_label}
S.~Lucas and T.~Reynolds, ``Learning deterministic finite automata with a smart
  state labeling evolutionary algorithm,'' \emph{IEEE transactions on pattern
  analysis and machine intelligence}, vol.~27, pp. 1063--74, Aug. 2005.

\bibitem{evo_dfa}
J.~G{\'o}mez, ``An incremental-evolutionary approach for learning deterministic
  finite automata,'' in \emph{2006 IEEE International Conference on
  Evolutionary Computation}.\hskip 1em plus 0.5em minus 0.4em\relax IEEE, 2006,
  pp. 362--369.

\bibitem{active_learning}
A.~Bartoli, A.~De~Lorenzo, E.~Medvet, and F.~Tarlao, ``Active learning
  approaches for learning regular expressions with genetic programming,'' in
  \emph{Proceedings of the 31st Annual ACM Symposium on Applied Computing},
  2016, pp. 97--102.

\bibitem{rosenberg_mfa}
A.~Rosenberg, ``On multi-head finite automata,'' \emph{IBM Journal of Research
  and Development}, vol.~10, pp. 388--394, Sep. 1966.

\bibitem{mfa}
M.~Holzer, M.~Kutrib, and A.~Malcher, ``Multi-head finite automata:
  Characterizations, concepts and open problems,'' \emph{Electronic Proceedings
  in Theoretical Computer Science}, vol.~1, p. 93–107, Jun. 2009.

\bibitem{atari}
V.~Mnih \emph{et~al.}, ``Playing atari with deep reinforcement learning,''
  2013.

\bibitem{target_net}
------, ``Human-level control through deep reinforcement learning,''
  \emph{Nature}, vol. 518, no. 7540, p. 529, 2015.

\bibitem{genetic}
A.~Thengade and R.~Dondal, ``Genetic algorithm-survey paper,'' in \emph{MPGI
  National Multi Conference}.\hskip 1em plus 0.5em minus 0.4em\relax Citeseer,
  2012, pp. 7--8.

\bibitem{transducer}
A.~Esmoris, C.~I. Chesñevar, and M.~P. González, ``Tags: A software tool for
  simulating transducer automata,'' \emph{The International Journal of
  Electrical Engineering \& Education}, vol.~42, no.~4, pp. 338--349, 2005.

\end{thebibliography}

\end{document}